# Classification of Operational Records in Aviation Using Deep Learning Approaches

Aziida Nanyonga
School of Engineering and Information Technology
University of New South Wales
Canberra, Australia
a.nanyonga@adfa.edu.au

Graham Wild
School of Science
University of New South Wales
Canberra, Australia
g.wild@adfa.edu.au

*Abstract*—Ensuring safety in the aviation industry is critical, even minor anomalies can lead to severe consequences. This study evaluates the performance of four different models for DP (deep learning), including Bidirectional Long Short-Term Memory (BLSTM), Convolutional Neural Networks (CNN), Long Short-Term Memory (LSTM), and Simple Recurrent Neural Networks (sRNN), on a multi-class classification task involving Commercial, Military, and Private categories using the Socrata aviation dataset of 4,864 records. The models were assessed using a classification report, confusion matrix analysis, accuracy metrics, validation loss and accuracy curves. Among the models, BLSTM achieved the highest overall accuracy of 72%, demonstrating superior performance in stability and balanced classification, while LSTM followed closely with 71%, excelling in recall for the Commercial class. CNN and sRNN exhibited 67% and 69% lower accuracies, with significant misclassifications in the Private class. While the results highlight the strengths of BLSTM and LSTM in handling sequential dependencies and complex classification tasks, all models faced challenges with class imbalance, particularly in predicting the Military and Private categories. Addressing these limitations through data augmentation, advanced feature engineering, and ensemble learning techniques could enhance classification accuracy and robustness. This study underscores the importance of selecting appropriate architectures for domain-specific tasks and contributes to advancing deep learning applications in multi-class classification problems

## I. Introduction

The aviation industry generates vast amounts of operational data through daily flight operations, including maintenance logs, flight operation summaries, incident reports, and other safety-related documentation [1]. These records are essential for identifying trends, improving operational efficiency, and ensuring aviation safety across various sectors, including military, private, and commercial aviation [2]. However, with the increasing volume and diversity, manual classification and analysis methods have become impractical and time-consuming [3, 4]. Furthermore, the unstructured and diverse nature of operational narratives poses significant challenges for efficient classification and analysis. Automating these processes is essential for identifying safety risks, operational anomalies, and issues that may compromise aviation safety [5, 6]. Misclassifications in this context could delay responses to safety concerns or lead to inaccurate trend analysis, ultimately jeopardizing operational efficiency and safety [7].

Traditional approaches, such as rule-based systems and manual reviews, often fail to fully capture the nuanced, context-dependent information embedded in text-based records. With the exponential growth of data, their limitations have become more pronounced [8]. Modern Deep Learning (DL) and Natural Language Processing (NLP) techniques offer a viable alternative, having proven highly effective in automating complex classification tasks in various domains [9]. DL models such as Bidirectional LSTMs (BLSTMs), Convolutional Neural Networks (CNNs), Long Short-Term Memory (LSTM), and Recurrent Neural Networks (RNNs), networks, have shown remarkable performance in sequential data tasks like text classification, sentiment analysis, and speech recognition [10].

Aviation operational records require models capable of understanding context, sequence, and long-range dependencies. RNNs [11], designed for sequential data, are commonly used in text classification tasks [12, 13] but face challenges with long-term dependencies caused by "exploding or vanishing gradients". LSTM network [14] address these limitations through mechanisms that retain and manage information over longer sequences. BLSTM further enhances performance by capturing context from both forward and backward directions, making them particularly effective for tasks requiring sequence sensitivity. Additionally, CNNs, traditionally applied to image processing, have shown strong capabilities in text classification by capturing local features and patterns in textual data [15].

Despite these advances, the application of DL to aviation safety records has been relatively underexplored. Most research in this domain focuses on traditional rule-based methods, leaving the potential of DL largely untapped. This study aims to fill this gap by leveraging CNN, RNN, LSTM, and BLSTM models to classify aviation operational records from the Socrata dataset into three categories: military, private, and commercial sectors. The Socrata dataset, with its extensive collection of public aviation records, offers a valuable resource for evaluating these models. By applying these advanced models, this research explores how DL can enhance the automation and accuracy of classifying operational records, contributing to improved aviation safety. Accurate classification facilitates efficient risk assessment, regulatory compliance, and the timely identification of safety concerns, all of which are crucial for reducing incidents and enhancing safety standards in aviation.

This study provides a novel application of DL for classifying aviation operational records and contributes to the growing body of research on NLP and DL in aviation safety. The findings underscore the practical benefits of DL in this field and offer insights into improving automated systems for operational data classification across various aviation sectors.

The structure of this paper is as follows: the next section reviews related work, highlighting existing approaches and advancements in the domain, followed by a detailed methodology section that outlines the data preprocessing steps, model architecture, and evaluation metrics employed in this study. The results are then presented, followed by a discussion of the findings, providing an in-depth analysis of model performance and their implications. The paper concludes with a conclusion and directions for future research aimed at improving the classification of aviation operational records.

## II. RELATED WORK

The classification of operational records in aviation, particularly those related to safety, has garnered significant interest among researchers over the years. Traditionally, aviation safety data such as incident reports, maintenance logs, and flight operation summaries have been analyzed using rule-based systems, expert systems, and manual review methods. While these approaches have been effective to some extent, they are increasingly criticized for being time-consuming and incapable of addressing the complexities of the growing volume and diversity of operational records [16, 17].

Historically, rule-based classification techniques have been employed to analyze aviation incident reports. These methods depend on predefined rules, such as keyword matching and heuristic approaches, to categorize records into specific classes. For instance, researchers have used decision trees (DT) and Naive Bayes (NB) [18], as well as random forest (RF), support vector machines (SVM), and NN [19], to be able to classify incident reports based on predefined features. However, the inherent limitation of these methods lies in their inability to understand the contextual nuances within text data, which is crucial for accurate classification.

Additionally, traditional machine learning algorithms such as Logistic Regression (LR) as well as Random Forests (RF) have been applied to predict the severity of aviation incidents. For example, Inan et al. [20] demonstrated the application of models such as LR, artificial neural networks (ANN), and DTs to classify aircraft damage based on various factors, including zones, weather, and incident severity. Although these models showed promise, they required extensive feature engineering and struggled to identify complex patterns within unstructured text data.

With advancements in DL, researchers have increasingly adopted more sophisticated methods to address the limitations of traditional approaches. CNN, RNN, LSTM and BLSTM [21] models have proven particularly effective for handling aviation safety data. These models excel in capturing complex patterns and dependencies, making them ideal for tasks like classifying aviation safety reports, which often involve textual or sequential information [22].

Several studies have highlighted the utility of DL in aviation safety analysis. For instance, Zhang et al., applied LSTM networks to classify aviation safety reports, leveraging the sequential nature of the data to improve prediction accuracy. Their results indicated that DL models outperformed traditional machine learning algorithms in classifying incident severity and conducting risk assessments [23].

Another significant study focused on classifying flight phases in safety reports from the Australian Transport Safety Bureau (ATSB). This study employed NLP and DL models, including CNN, LSTM, BLSTM, and simple RNN. The models were evaluated based on usual metrics; specifically, their accuracy, precision, recall, and F1 score. Among these, LSTM achieved the highest performance, with accuracy, precision, recall, and F1 scores of 87%, 88%, 87%, and 88%, respectively. These findings underscore the effectiveness of DL models in automating the analysis of safety occurrences, enabling more precise safety measures and efficient report handling [21].

Recent efforts have also explored combining DL models for improved performance [24, 25]. These studies used DL to try and classify the phase of flight in which an aviation safety occurrence was reported, by utilizing the information provided in the associated text narratives. These efforts demonstrated almost 90% accuracy.

Despite these advances, gaps remain in the literature regarding comparative analyses of DL models for aviation safety classification tasks. While numerous studies have explored CNN, LSTM, BLSTM, and RNN models on various datasets, there is limited research on direct comparisons of their effectiveness in classifying operational records across sectors such as military, private, and commercial aviation. Furthermore, the application of DL to the Socrata dataset for aviation operational record classification is largely unexplored.

This study addresses these gaps by evaluating and comparing the performance of CNN, RNN, LSTM, and BLSTM models in classifying aviation operational records. Leveraging the Socrata dataset, this research aims to analyze each model's strengths and limitations. The findings will offer insights into the potential of DL for improving aviation safety analysis, operational efficiency, and data classification

## III. METHODOLOGY

The methodology employed in this study was to classify operator types (military, private, and commercial) using narratives from the Socrata aviation dataset. The methodology is divided into several key subsections: Data Annotation and Preprocessing, DL Model Architecture, and Model Performance Evaluation.

### A. Data Annotation and Preprocessing

The Socrata aviation dataset, which is publicly available online, serves as the primary data source for this study. The original dataset contained 4,995 records, with each record consisting of an operator type and a narrative describing aviation incidents. After cleaning the data, the dataset was reduced to 4,863 records. The focus of this study was on the *Operator* and narratives labelled as *Summary* fields. The Operator field represents the entity responsible for the operation of the aircraft, while the Narrative field provides a descriptive account of the incident.

The operators in the dataset were originally labelled with hundreds of categories. To streamline the classification process and ensure meaningful analysis, these categories were manually annotated and grouped into three primary classes: *Military*, *Private*, and *Commercial*. This annotation was performed by cross-checking each operator category with additional data sources and domain knowledge. This manual verification process helped to ensure that the categories were consistent and representative of the operators in the dataset. The categorization of operators into these three distinct classes allowed for a focused and meaningful classification task, ensuring that the models trained on this data could accurately predict operator types based on the narrative content.

### B. Text Processing

Text preprocessing is needed to transform the raw text data from the reports into a format suitable for use in training the models. The text preprocessing pipeline in this study involved several stages, including text cleansing, tokenization, and word count distribution analysis.

The first step in text preprocessing was text cleansing, where irrelevant elements such as stop words (e.g., "the", "and" "is"), punctuation and special characters were removed. This was done using popular NLP libraries, such as *NLTK* and *SpaCy*. The removal of stop words is particularly important because these words do not add significant meaning to the classification task but can introduce noise into the model, reducing its performance [26].

Following text cleansing, tokenization was performed using the Keras Tokenizer. Tokenization involves splitting each narrative into individual words or tokens and assigning each token a unique integer ID. This step is essential for converting text into a numerical format that DL models can process. The tokenized text was then padded to a fixed length of 200 words, ensuring that all input sequences had the same length. Narratives shorter than 200 words were padded with zeros, while longer ones were truncated to maintain consistency in sequence length across the dataset.

An analysis of the word count distribution within the Socrata dataset revealed that the narratives varied greatly in length. While some narratives were quite short, others contained hundreds of words. As seen in Fig. 1, this variance prompted the decision to standardize the input sequence length for the DL models. By ensuring that the models received consistent input, the approach aimed to optimize model performance and avoid issues associated with input length inconsistency.

Finally, the dataset was split into three parts for training, testing, and validation. The data splitting utilised the standard 80-10-10 ratio, where 80% of the data being utilised for model training, 10% for testing, and the remaining 10% for validation. This split is needed to be able to evaluate model performance utilizing unseen data helping to prevent overfitting the model based only on the training data.

*C. DL Model Architecture*

To ensure consistency and enable effective comparisons across all models, a unified architecture was used as the base, with slight modifications tailored to each model. The architecture consisted of three main components; the first is the embedding layer, then the hidden layers, and the output layer. Within hidden layers, activation utilized a Rectified Linear Unit (ReLU) function, which allowed the model to capture any complex, non-linear relationships. Within the output layer, the activation function utilized was SoftMax, enabling its use for multi-class classification. This approach allowed the model to output a probability distribution across the possible classes. The resultant class prediction utilized an argmax function to identify corresponding the index with the highest probability from the output of SoftMax function. A diagram illustrating the DL architectures used in this study is provided in Fig. 1.

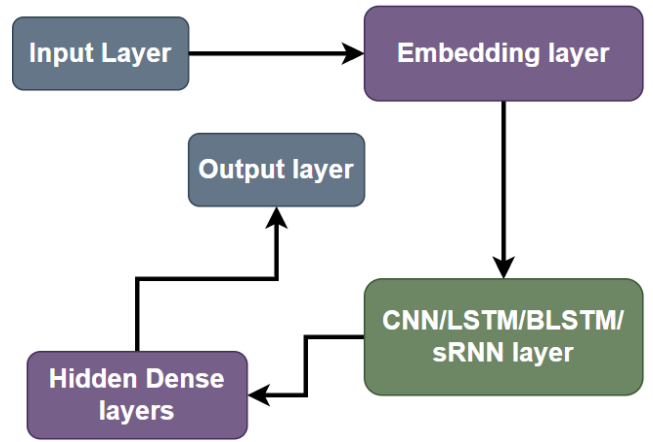

Fig. 1. Methodology Archtecture

*1) sRNNs.*

The Simple RNN (sRNN) is a basic form of RNN designed to process sequential data by using the output from the previous time step as input for the current time step. Its architecture is straightforward, typically comprising a single hidden layer that enables the flow of information across successive time steps [11]. The mathematical formulation for an sRNN can be expressed as follows:

$$h_t = \sigma(W_h \cdot [h_{t-1}, x_t] + b_h), \qquad (1)$$

where $h_t$ donates the hidden state at the time step *t*, while $x_t$ is the input at the corresponding time step, and $\sigma$ is the activation function used.

*2) LSTM*

The LSTM network is an advanced variant of the RNN that was designed to overcome the challenge of capturing long-term dependencies that may be present within sequential data. In contrast to basic RNNs, LSTMs are equipped with memory cells that allow for the storage and management of information over extended periods; this makes them particularly effective when tasks need an understanding of context and temporal patterns. The LSTM architecture is more intricate than that of a standard RNN, incorporating three gates, the forget, input, and output gates [14]. These gates collectively manage information flow, determining what data should be retained and what should be discarded.

*3) CNN*

The CNN is a DL model primarily used for image processing but is also effective in sequential data tasks like text classification. The CNN is made up of convolutional layers that utilize filters to analyze input data, detecting localized features and patterns. In text processing, CNNs help identify important word-level features, such as n-grams, which are crucial for understanding context [27].

*4) BLSTM*

The BLSTM model extends the traditional LSTM architecture by processing input sequences in both forward and backward directions. This bidirectional approach enables the model to capture information from both past and future contexts, thereby enhancing its ability to comprehend dependencies across the entire sequence. Such a feature makes BLSTM particularly effective in tasks requiring a holistic understanding of sequence context. Mathematically, the BLSTM model combines the operations of both forward and backward LSTMs, allowing for a more robust capture of long-term dependencies [22]. BLSTM is widely applied in domains such as speech recognition, language translation, and named entity recognition, where the integration of both past and future information is crucial for precise prediction and analysis

*D. Some Common Mistakes*

This section elucidates the evaluation criteria utilized in this study to assess the models' performance. The primary focus of this research is multi-class classification, and as such, performance was gauged based on the accuracy of predictions across various classes. To comprehensively evaluate model performance, we employed a suite of standard prediction performance metrics; specifically, recall, F1-score, precision, and accuracy.

## IV. RESULTS AND DISCUSSION

The evaluation of the four models CNN, sRNN, BLSTM, and LSTM on the given dataset, incorporating both training and validation metrics, is presented. The performance of each model is assessed through classification reports, confusion matrices, accuracy, macro averages, as well as validation loss and accuracy

*A. Model Performance: Classification Report*

The classification report evaluates the models' precision, recall, and F1 scores across three classes: Commercial, Military, and Private. Among the models, CNN excelled in classifying the Commercial class, achieving 81% precision and 79% recall. However, its performance dropped significantly for the Military and Private classes, with much lower recall and precision values. Similarly, the sRNN model achieved the highest recall (85%) for the Commercial class but failed to classify the Private class altogether, resulting in zero precision, recall, and F1-score for this category.

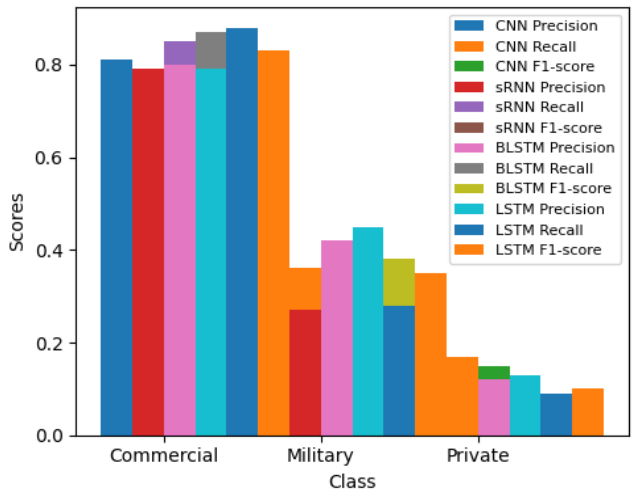

Fig. 2. Precision, Recall, F1-Score Comparison Across Models

*B. Confusion Matrix Analysis*

The confusion matrices (Fig. 4) provide insights into the class-wise predictions and misclassifications. The CNN model demonstrated strong predictive accuracy for the Commercial class but showed higher misclassifications for the Military and Private classes, reflecting its challenges in distinguishing these categories. The sRNN model's confusion matrix highlights significant misclassification of the Private class, aligning with its low scores in the classification report.

On the other hand, BLSTM's confusion matrix showed relatively fewer misclassifications, particularly for the Commercial and Military classes, indicating its ability to capture complex dependencies within the data. LSTM also performed well for the Commercial class but exhibited a moderate level of misclassification for the Private class, performing better than sRNN but worse than BLSTM. These trends emphasize the models' varying capabilities to manage class imbalances and complexities in the dataset.

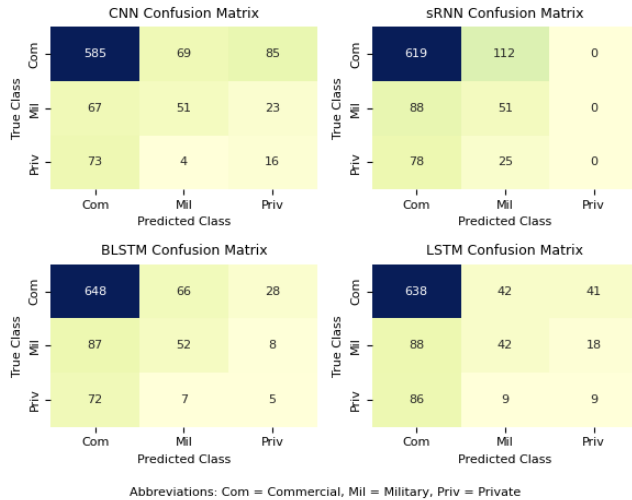

Fig. 3. Confusion Matrix for all of the four models

*C. Model Accuracy Comparison*

The overall performance of each model, summarized in Table I, highlights BLSTM's superior accuracy of 72%, closely followed by LSTM at 71%. These results demonstrate the ability of these models to correctly classify instances across classes. In contrast, sRNN and CNN achieved lower accuracies of 69% and 67%, respectively, reflecting their limitations in handling the Private and Military classes.

Notably, BLSTM and LSTM's ability to incorporate temporal dependencies likely contributed to their superior performance, as this approach enables better context understanding across sequences. CNN's comparatively lower accuracy suggests its limitations in capturing sequential relationships, while sRNN's architecture may have struggled with the dataset's complexity.

TABLE I. DEEP LEARNING MODEL PERFORMANCE

| Models | Precision | Recall | F1 Score | Accuracy |
|---|---|---|---|---|
| **CNN** | 0.67 | 0.67 | 0.68 | 0.67 |
| **sRNN** | 0.69 | 0.69 | 0.66 | 0.69 |
| **LSTM** | **0.72** | **0.72** | **0.70** | **0.72** |
| **BLSTM** | 0.71 | 0.71 | 0.68 | 0.71 |

*D. Macro Average Comparison*

Macro average precision, recall, and F1-scores, as shown in Fig. 5, provide a balanced evaluation of the models across all classes, irrespective of class imbalances. BLSTM and LSTM outperformed CNN and sRNN in all three metrics, particularly excelling in recall for the Commercial class. However, all models showed reduced performance for the Private class, highlighting a consistent challenge in handling this category.

This underperformance could be attributed to class imbalance or insufficient distinguishing features within the dataset for the Private class. Addressing these challenges through techniques like data augmentation, oversampling, or feature engineering could improve overall performance.

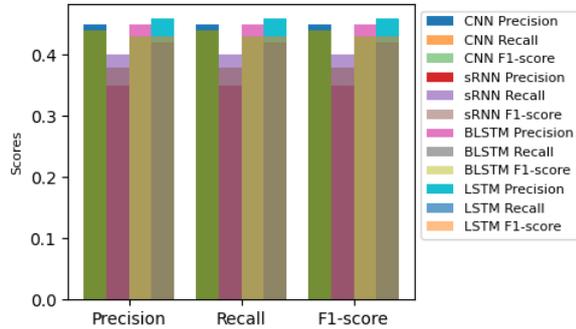

Fig. 4. Macro Average Comparison for all Models

*E. Validation metrics*

The validation loss and accuracy trends (Fig. 6) reveal each model's generalization capabilities. BLSTM exhibited the lowest validation loss and the highest validation accuracy, reflecting its stable learning process and strong generalization to unseen data. LSTM followed closely, showing similar trends with slightly higher validation loss.

Conversely, CNN and sRNN showed higher validation losses, indicating weaker generalization and potential overfitting to the training data. These results underscore the importance of architectures that can effectively manage sequential dependencies, as demonstrated by BLSTM and LSTM.

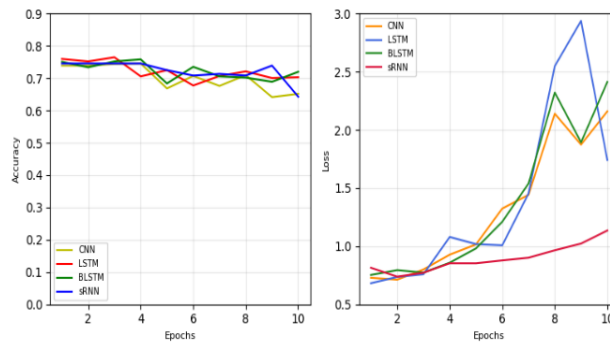

Fig. 5. Validation Accuracy (right) and Validation loss (left) for all Models

## V. CONCLUSION

This research employed four DL models CNN, sRNN, BLSTM, and LSTM to evaluate their performance in a multi-class classification task. The findings highlight the strengths and weaknesses of each of the models. CNN achieved an accuracy of 67%, displaying a balanced performance but facing challenges in classifying the Military and Private classes, as reflected in lower recall and F1 scores for these categories. sRNN performed slightly better with an accuracy of 69%, although it struggled particularly with the Private class, where recall and F1 scores were negligible.

BLSTM emerged as the most effective model, achieving the highest accuracy (72%) and demonstrating balanced classification across all classes, particularly excelling in recall for the Commercial class. LSTM followed closely with an accuracy of 71%, showing strong recall for the Commercial class but encountering similar difficulties with the Military and Private classes. Validation loss and accuracy metrics further corroborated these results, with BLSTM and LSTM demonstrating lower validation loss and greater stability, indicating their superior generalization ability.

Despite the promising results, the study revealed key challenges, particularly in accurately classifying the Military and Private classes. These challenges likely stem from class imbalances and limited feature representation in the dataset. Future improvements could include the use of data augmentation and oversampling techniques to address class imbalance, the integration of attention mechanisms to enhance model focus on underrepresented categories, and the application of ensemble learning to leverage the strengths of multiple models. Additionally, domain-specific feature engineering could play a crucial role in improving classification performance for the minority classes.

In conclusion, while BLSTM proved to be the most effective model for this classification task, further refinements and advanced techniques are necessary to overcome the observed limitations, paving the way for more robust performance in multi-class classification problems.

REFERENCES

[1] G. S. Baxter and G. Wild, "Aviation Safety, Freight, and Dangerous Goods Transport by Air," in *International Encyclopedia of Transportation*, R. Vickerman, Ed. Oxford: Elsevier, 2021, pp. 98-107.
[2] A. Ayiei, J. Murray, and G. Wild, "Visual Flight into Instrument Meteorological Condition: A Post Accident Analysis," *Safety*, vol. 6, no. 2, pp. 1-25, 2020, Art. no. 19.
[3] A. Nanyonga, H. Wasswa, U. Turhan, K. Joiner, and G. Wild, "Comparative Analysis of Topic Modeling Techniques on ATSB Text Narratives Using Natural Language Processing," in *2024 3rd International Conference for Innovation in Technology (INOCON)*, 2024, pp. 1-7: IEEE.
[4] A. Nanyonga, H. Wasswa, and G. Wild, "Topic Modeling Analysis of Aviation Accident Reports: A Comparative Study between LDA and NMF Models," in *2023 3rd International Conference on Smart Generation Computing, Communication and Networking (SMART GENCON)*, 2023, pp. 1-2: IEEE.
[5] J. Liu, H. Yan, and Y. Du, "Application of Text Analysis Technology in Aviation Safety Information Analysis," *Journal of Physics: Conference Series*, vol. 1624, no. 3, p. 032033, 2020/10/01 2020.
[6] A. Nanyonga, H. Wasswa, U. Turhan, K. Joiner, and G. Wild, "Exploring Aviation Incident Narratives Using Topic Modeling and Clustering Techniques," in *2024 IEEE Region 10 Symposium (TENSYMP)*, 2024, pp. 1-6: IEEE.
[7] K. Goebel, N. Eklund, and B. Brunell, "Rapid detection of faults for safety critical aircraft operation," in *2004 IEEE Aerospace Conference Proceedings (IEEE Cat. No. 04TH8720)*, 2004, vol. 5, pp. 3372-3383: IEEE.
[8] K. Darveau, D. Hannon, and C. Foster, "A comparison of rule-based and machine learning models for classification of human factors aviation safety event reports," in *Proceedings of the Human Factors and Ergonomics Society Annual Meeting*, 2020, vol. 64, no. 1, pp. 129-133: SAGE Publications Sage CA: Los Angeles, CA.
[9] A. Nanyonga and G. Wild, "Impact of Dataset Size & Data Source on Aviation Safety Incident Prediction Models with Natural Language Processing," in *2023 Global Conference on Information Technologies and Communications (GCITC)*, 2023, pp. 1-7: IEEE.
[10] !!! INVALID CITATION !!! [6, 10, 11].
[11] S. Grossberg, "Recurrent neural networks," *Scholarpedia*, vol. 8, no. 2, p. 1888, 2013.
[12] A. Nanyonga, H. Wasswa, and G. Wild, "Comparative Study of Deep Learning Architectures for Textual Damage Level Classification," in *2024 11th International Conference on Signal Processing and Integrated Networks (SPIN)*, 2024, pp. 421-426: IEEE.

[13] S. M. Zaman, M. M. Hasan, R. I. Sakline, D. Das, and M. A. Alam, "A comparative analysis of optimizers in recurrent neural networks for text classification," in *2021 IEEE Asia-Pacific Conference on Computer Science and Data Engineering (CSDE)*, 2021, pp. 1-6: IEEE.
[14] S. Hochreiter, "Long Short-term Memory," *Neural Computation MIT-Press,* 1997.
[15] J. Zhang, Y. Li, J. Tian, and T. Li, "LSTM-CNN Hybrid Model for Text Classification," in *2018 IEEE 3rd Advanced Information Technology, Electronic and Automation Control Conference (IAEAC)*, 2018, pp. 1675-1680.
[16] F. Gürbüz, L. Özbakir, and H. Yapici, "Classification rule discovery for the aviation incidents resulted in fatality," *Knowledge-Based Systems,* vol. 22, no. 8, pp. 622-632, 2009/12/01/ 2009.
[17] A. Nanyonga, K. Joiner, U. Turhan, and G. Wild, "Applications of natural language processing in aviation safety: A review and qualitative analysis," in *AIAA SCITECH 2025 Forum*, 2025, p. 2153.
[18] E. Kuşkapan, M. A. Sahraei, and M. Y. Çodur, "Classification of Aviation Accidents Using Data Mining Algorithms," (in en), *Balkan Journal of Electrical and Computer Engineering,* vol. 10, no. 1, pp. 10-15, January 2022.
[19] I. Rosadi, F. Franciscus, and M. H. Widanto, "A Machine Learning-Based Method for Predicting the Classification of Aircraft Damage," *EAI Endorsed Transactions on Internet of Things,* vol. 10, 08/15 2024.
[20] T. T. İnan, "Aircraft Damage Classification by using Machine Learning Methods," *International Journal of Aviation, Aeronautics, and Aerospace,* vol. 10, no. 2, p. 4, 2023.
[21] A. Nanyonga, H. Wasswa, and G. Wild, "Aviation Safety Enhancement via NLP & Deep Learning: Classifying Flight Phases in ATSB Safety Reports," in *2023 Global Conference on Information Technologies and Communications (GCITC)*, 2023, pp. 1-5: IEEE.
[22] A. Nanyonga, H. Wasswa, O. Molloy, U. Turhan, and G. Wild, "Natural language processing and deep learning models to classify phase of flight in aviation safety occurrences," in *2023 IEEE Region 10 Symposium (TENSYMP)*, 2023, pp. 1-6: IEEE.
[23] X. Zhang, P. Srinivasan, and S. Mahadevan, "Sequential deep learning from NTSB reports for aviation safety prognosis," *Safety science,* vol. 142, p. 105390, 2021.
[24] A. Nanyonga, H. Wasswa, and G. Wild, "Phase of Flight Classification in Aviation Safety Using LSTM, GRU, and BiLSTM: A Case Study with ASN Dataset," in *2023 International Conference on High Performance Big Data and Intelligent Systems (HDIS)*, 2023, pp. 24-28: IEEE.
[25] A. Nanyonga, H. Wasswa, U. Turhan, O. Molloy, and G. Wild, "Sequential Classification of Aviation Safety Occurrences with Natural Language Processing," in *AIAA AVIATION 2023 Forum*, 2023, p. 4325.
[26] S. Popova, L. Kovriguina, D. Mouromtsev, and I. Khodyrev, "Stop-words in keyphrase extraction problem," in *14th Conference of Open Innovation Association FRUCT*, 2013, pp. 113-121.
[27] K. O'Shea, "An introduction to convolutional neural networks," *arXiv preprint* 2015.